\RequirePackage{amsmath}
\documentclass[runningheads]{llncs}
\usepackage{amsmath}
\usepackage{graphicx}
\usepackage{float} 
\usepackage[inline]{enumitem}
\usepackage{scrextend}
\usepackage{appendix}

\addtokomafont{labelinglabel}{\sffamily\bfseries}

\graphicspath{{figures/}}

\begin{document}
\title{GarmNet: Improving Global with Local Perception for Robotic Laundry Folding}
\titlerunning {GarmNet: Improving Global with Local Perception for Robotic Laundry...}

%
%
\author{Daniel Fernandes Gomes\inst{1}\orcidID{0000-0001-8619-4966} \and
Shan Luo\inst{1}\orcidID{0000-0003-4760-0372} \and
Luis F. Teixeira\inst{2,3}\orcidID{0000-0002-4050-7880}}
%
%
\institute{Department of Computer Science,\\ 
University of Liverpool, Liverpool, UK\\
\email{\{danfergo,shan.luo\}@liverpool.ac.uk}\\
\and
Faculdade de Engenharia, Universidade do Porto, Portugal \\
\and
INESC TEC, Porto, Portugal \\
\email{luisft@fe.up.pt}}
\maketitle              
\sloppy 
\begin{abstract}
 Developing autonomous assistants to help with domestic tasks is a vital topic in robotics research. Among these tasks, garment folding is one of them that is still far from being achieved mainly due to the large number of possible configurations that a crumpled piece of clothing may exhibit. Research has been done on either estimating the pose of the garment as a whole or detecting the landmarks for grasping separately. However, such works constrain the capability of the robots to perceive the states of the garment by limiting the representations for one single task. In this paper, we propose a novel end-to-end deep learning model named \textit{GarmNet} that is able to simultaneously localize the garment and detect landmarks for grasping. The localization of the garment represents the global information for recognising the category of the garment, whereas the detection of landmarks can facilitate subsequent grasping actions. We train and evaluate our proposed GarmNet model using the CloPeMa Garment dataset that contains 3,330 images of different garment types in different poses. The experiments show that the inclusion of landmark detection (GarmNet-B) can largely improve the garment localization, with an error rate of 24.7\% lower. Solutions as ours are important for robotics applications, as these offer scalable to many classes, memory and processing efficient solutions.

\end{abstract} 
\section{Introduction} \label{section:intro}

Garment recognition is one necessary capability not only for the automation of
tasks involving its manipulation, such as garment folding, within robotic systems  \cite{pr2Towels} but many other applications as well: online e-commerce platforms \cite{surveillance} that make suggestions based on image information \cite{liu2016deepfashion}, intelligent surveillance systems that track people based on the clothing description, etc. 
However, recognition of clothes, or (highly) deformable objects in general, is a challenging task due to the many poses and deformations that a flexible object may exhibit.



We consider the scenario wherein an image containing a single piece of clothing, flat, wrinkled and semi-folded exists on a clean background and a robotic system wants to find the garment and good grasping points. Our goal is then to perceive, at the global level, a piece of garment existing in an image, by localizing and classifying it. And, at a local level, by identifying and localizing its landmarks, e.g.,  neckline-left, right-armpit, right-sleeve-inner, is an image point. Each garment class has a different amount and types of landmarks e.g., towels have four, whereas t-shirts and long t-shirts have both 12. Because of such variances, garment+landmark detection can be formulated using two different approaches: \begin{enumerate*}[label=(\alph*)] \item \label{item:slc_a} garment finding as object localization, followed by conditional, class specific, landmarks finding, also as object localization. \item Finding all landmarks existing in the image independently of the garment class as object detection, and the garment piece as object location. \end{enumerate*}

Although approach \ref{item:slc_a}, is a simpler solution to build using off-the-shelf models, and is commonly seen in current literature \cite{Li_ICRA2014} \cite{Mariolis2015}, it is more inefficient because: \begin{enumerate*}\item It requires one different sub-model for each garment category, being that many landmarks are shared between garment category e.g., a sleeve of a hoody is similar to a sleeve in a jacket. \item  On the other hand, when using Neural Networks, these work by building more complex representations as its depth increases, through the combinatorial effect of chaining multiple layers together. Therefore, a Network that recognizes a piece of clothing, should in principle, recognize somewhere in its hidden layers some of its landmarks. \end{enumerate*} This means that in approach \ref{item:slc_a} multiple redundant hidden features are learned, representing extra parameters to be stored in memory, and more operations to be computed during execution time. Adding to this, in the context of robotics e.g., a top view of a laundry bin, the garment global perception might not be possible with good accuracy, but recognizing some local landmarks might be just as valuable for the robot, as it could grasp the piece of clothing by one good landmark and then perform further recognition using the same model.
    
Therefore, we  address the detection of landmarks and classification+localization of garment simultaneously, with a Convolutional Neural Network (CNN) \cite{yannLecunConvNets} composed of one common trunk and two separate branches. We then introduce a bridge connection that feeds the landmarks detection output into the garment localizer branch, resulting in a decrease from 56.7\% to 32.0\% in the error rate, that demonstrates the advantages in considering both tasks together. We balance and augment our dataset with Gaussian and hue noise, and perform one last training achieving: 0\% and 17.8\% error rate on classification and classification+localization respectively; and 36.2\% mean Average Precision (mAP) on landmark detection.

\section{Related Work}

\paragraph{Early works} handling clothes happen in Robotics with the folding task, and the problem domain is constrained enough to avoid the necessity of classification i.e., only one type of clothing is considered. In \cite{pr2Towels} towels are considered, and depth discontinuities are explored to detect its borders and corners. With that information, a PR2 robot is able pick them from a random dropped position and, following a predefined sequence of steps, folds them.

\paragraph{Machine Learning methods} are latter used, not only in robotic tasks but other software applications as well e.g., in \cite{surveillance} real-time classification and segmentation of people's clothing appearing in a camera video feed are addressed. However, using the raw image i.e., each pixel is a feature, would result in a low performance results due to the  curse of dimensionality, that many Machine Learning (ML) methods suffer from. To overcome this challenge, one common approach is to use the Bag of Visual Words (BoW), that extracts handcrafted features e.g., Scale-Invariant Feature Transform (SIFT) or Histogram of Oriented Gradients (HOG) and feeds these into a classifier e.g., Support Vector Machine (SVM) or k-Nearest Neighbours (k-NN). In \cite{Li_ICRA2014} the authors use this approach to design a two layer hierarchical classifier to determine the category and grasping point (equivalent to pose, in this work) of a piece of clothing. 
Other works address the extraction of domain specific features from images. In \cite{Yamazaki2016}, a set of Gabor filters are used to filter edges magnitude and orientation that are representative of wrinkling and overlaps and with that information, the authors propose three types of features: Position and orientation distribution density, cloth fabric and wrinkle density; and existence of cloth-overlaps.

\paragraph{Deep Learning (DL) methods.} In 2012, AlexNet \cite{Krizhevsky_imagenetclassification} achieves a notorious improvement of 11\% in the  ILSVRC2012 image classification competition, when comparing against the next best solution. 
In \cite{r-cnn_0}, to address object detection, region proposal techniques are combined with CNNs, resulting in R-CNN, an hybrid method that combines a CNN with SVMs. Then, in \cite{fast_rcnn}, two main improvements are made: the RoI pooling layer is introduced, and the SVM  is replaced by a softmax classifier. The improved model Fast R-CNN model is then a two headed CNN, optimized using a multi-task loss, $9 \times$ quicker to train and $213 \times$ faster test-time. 
In \cite{faster_rcnn}, the authors further introduce Region Proposal Networks (RPNs), making this architecture finally an end-to-end trainable model. Another simpler architecture, YOLO, is introduced in \cite{yolo} and improved in \cite{yolo9000}, consisting of only two direct branches on the top of a regular fully CNN: one for positions coordinates regression and another for classes attribution. The system is capable of real-time object detection.

\paragraph{DL in garment perception.} After the successes using DL methods, some works that address garment perception also explore their potential, mainly considering classification problems. One example is \cite{Mariolis2015}, that addresses the same problem as \cite{Li_ICRA2014}: pose and category recognition. The solution is also similar, a two layer hierarchical classifier. But here, instead of BoW and SVMs, one CNN is used to determine the garment class, followed by a class specific CNN that determines the garment pose index. 
The authors compare the model against others using hand engineered features and report gains of 3\% in accuracy. Similarly, in \cite{activeGarmentRecognition} a robotic system composed of two arms and a RGB-D camera, uses an hierarchy of CNNs with three specialization levels. At a first step one CNN classifies the garment piece, and then two others are used to find the first and second grasping points. 

\paragraph{Our approach.} In contrast to these approaches, ours leverages the intrinsic characteristics of CNNs and the architecture patterns from both classification/localization and detection models to perform the global and local perception in one step with a single CNN. Our model can also exploit prior knowledge gained to detect landmarks, to enhance the global garment perception. Our approach is therefore more memory and processing efficient than hierarchical solutions presented above. This happens because, in our approach, lower level layers are shared between the Global and Local perception components, as discussed in section \ref{section:intro}. 
Exploring these intrinsic characteristics of CNNs, in \cite{hcnn}, the authors propose Hierarchical Cnvolutional Neural Networks (H-CNN), to address hierarchical classification. With coarse/top-level classifications being extracted from hidden layers while finer/more-specific classes being predicted by the latest layers of the network.

\section{Network architecture}

Our network, \textbf{GarmNet}, at a macro level can be summarized into three blocks: \textbf{Feature extractor}, \textbf{Landmark detector} and \textbf{Garment localizer}. 

\begin{figure}
  \begin{center}
    \leavevmode
    \includegraphics[width=0.7\textwidth]{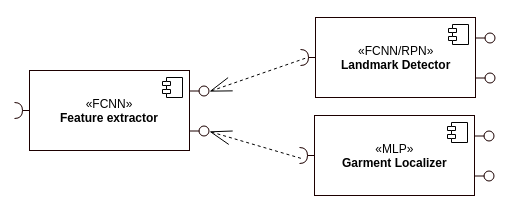}
    \caption{\textbf{GarmNet macro view,} UML \cite {uml} Components Diagram. The architecture is broken into three blocks (components): Feature extractor, Landmark Detector and Garment Localizer; that output: intermediate features at two depths, landmarks classes+localizations and garment class+localization. } 
    \label{fig:uml_macro_view}
  \end{center}
\end{figure}

\paragraph{Feature extractor} We implement the feature extraction module with a Fully Convolutional Neural Network (FCNN), a 50-layer ResNet\cite{resnet}. The model is pre trained on ImageNet\cite{imagenet_cvpr09}, to which we remove the last Fully Connected (FC) layers, resulting in a $7 \times 7$ output tensor. Yet, because in some cases we have multiple landmarks close to each other, we preferred a larger output size, that would result in a higher number of anchors in the landmark detector. We achieve this by probing the ResNet at the end of the \textit{conv4\_x} block, which has an output size of $14 \times 14$. 

\begin{figure}
  \begin{center}
    \leavevmode
    \includegraphics[width=0.5\textwidth]{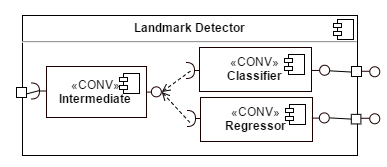}
    \caption{\textbf{Landmark detector} component, UML\cite{uml} representation. After one intermediate branch, two separate branches output $18\times18$ landmark proposals (classification and location). This block, implemented with convolutional layers, can be interpreted as small fully connected network sliding over the feature extractor output.} 
    \label{fig:arc_landmark_detector}
  \end{center}
\end{figure}

\paragraph{Landmark detector} Responsible for classifying and localizing all the landmarks present in the image, this module is a small, $3 \times 3$, sliding FCNN, similar to the Region Proposal Network (RPN) introduced in \cite{faster_rcnn}, and it is implemented with convolutional (CONV) layers. The network is composed by an intermediate 256-d layer, with a rectified linear unit (ReLU) \cite{Krizhevsky_imagenetclassification} activation, followed by two heads: one for localization and other for classification, see \ref{fig:arc_landmark_detector}. 

The localizer head is a Multi-Layer Perceptron (MLP), that outputs the predicted landmark relative coordinates such that
\begin{eqnarray}
	A_{(i,j)} = (s * j, s * i) \\
	L_{(i,j)} = A_{(i,j)} + O_{(i,j)}
\end{eqnarray}
where $L$ is a predicted landmark location, $A$ stands for the sliding network position $O$ the localization head output, and $s$ a stride value that we define to spread the base referential (or anchor points, as introduced in \cite{faster_rcnn}) evenly across the input image (we set it as 18, that together with a 26 landmark area matches the 224 image input size).
	
The classifier head, a convolutional layer with $n+1$ filters, where $n$ is the number of landmark classes and the plus one answers for background, non positive, landmarks. We apply a \textit{softmax} activation along the depth dimension and, therefore the output of this layer can be interpreted, at each position, as the probability of the associated landmark $L_{_(i,j)}$ being of a certain class, or background. 

\paragraph{Garment localizer} To perform the localization of the piece of clothing present in the image, we use a two head three-layer fully connected network, similar to the sliding window used for landmark detection. The intermediate layer is a 512-d FC with ReLU activation, followed by the regression and classification heads. The regression head outputs four values: x, y, with and height; and the classification head outputs $n$ values, where $n$ is the number of garment classes, that we remap to probabilities with a \textit{softmax} activation. We retrieve the predicted landmarks by computing the \textit{argmax} over the classifier output tensor depth dimension, and associating it with the point predicted in same spacial position on the regression head. We further discard all the landmarks that have a confidence value i.e., the value that motivated the argument to be the maximum, lower than 0.5. 

\begin{figure}
  \begin{center}
    \leavevmode
    \includegraphics[width=0.5\textwidth]{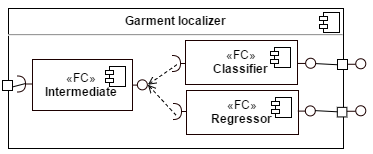}
    \caption{\textbf{Garment localizer component}, UML\cite{uml} representation.  Similar to the landmark detector \ref{fig:arc_landmark_detector}, yet fully connected layers are used. The Intermediate layer outputs a 512-d, the classifier a 9-d (one hot encoded classes) and the regressor the  3-d (x,y, with and height) vectors. }
    \label{fig:arc_garment_localizer}
  \end{center}
\end{figure}
\section{Experiments}

Our implementation was performed using Keras\footnote{https://keras.io/} framework with the TensorFlow\footnote{https://www.tensorflow.org/} back-end. All experiments were carried out on a laptop, with GPU support enabled, equiped with an Intel(R) Core(TM) i7-6700HQ CPU @ 2.60GHz, 16GB DDR4  RAM, and a Nvidia Geforce GTX 1060 6GB GDDR5 GPU. 
We initialize kernels with random values drawn from a normal distribution, and bias with ones. Optimization is performed using Adadelta with 1.0 learning, a batch size of 30, and 40 epochs per experiment. At test time our model runs at roughly 30 FPS. For classification and localization evaluation we use error rate, while for detection we use the mean Average Precision metric (mAP) as proposed in \cite{Everingham:2010:PVO:1747084.1747104}.
The source code has been made publicly available\footnote{ https://github.com/danfergo/garment}. 

\subsection{Dataset}
We adapt the CTU Color and Depth Image Dataset of Spread Garments \cite{ctuDataset}. This dataset is divided into two groups: ``Flat and wrinkled'',  with 2050,  and ``Folded'',  with 1280 examples. Each example contains one image of a piece of clothing placed on a wooden floor and is annotated with the stereo disparity; garment category; interest points coordinates and category; and other meta-information. We merge both groups, and  because it only contains information regarding each landmark position, we extend its annotation with the garment bounding box as follows:
\begin{eqnarray}
	\textstyle
	P_1 = (\min_{\forall l \in L} l_i, \min_{\forall l \in L} l_j) \\
    \textstyle
    P_2 = (\max_{\forall l \in L} l_i, \max_{\forall l \in L} l_j)
\end{eqnarray}
where $L$ is the set of Landmarks, $P_1$ and $P_2$ are the top-left and bottom-right corners of the bounding box. There is a total of 27 landmark categories, distributed among 9 types of garment.  Some landmark categories are shared among classes. We then create two splits, with 300 randomly chosen images for validation and the remaining  the ones used for training. The remaining 2318, make the training split. Results are reported over the validation split.

\subsection{Landmark detection anchors} \label{subsec:ground_truths}

For training the landmark detector heads we transform the landmark locations into small squared areas and follow a strategy similar to the anchor boxes described in \cite{faster_rcnn}. To all the anchor boxes that intercept a landmark box with $IoU > 0.7$, we consider it a positive for the respective class. If it does not intercept any with $IoU \ge 0.3$ we set it to background. Because the ratio between positive and negative anchor boxes is high, we create a binary mask that is used to filter the anchors that effectively are considered in the loss function. This mask selects all the class positive and 10 randomly chosen background anchors. 

\subsection{Loss functions}

\paragraph{Landmark detector} For the localization head, we apply the robust loss defined in \cite{faster_rcnn} to all the active anchors that are selectable by the mask, $m$, described in \ref{subsec:ground_truths}.

\begin{eqnarray}
	L_{reg} & = & m \odot L_{robust}(P, T)\\
	L_{robust}(P, T) & = & \left.
  \begin{cases}
    (P-T)^2, & P - T < 0.5 \\
    |P-T|, & \text{otherwise }
  \end{cases}
  \right.
\end{eqnarray}

For the classification head, we apply the cross-entropy loss function to all active anchors, being that at each anchor the landmark class is one-hot encoded.
\begin{eqnarray}
	L_{cls} = m \odot L_{ce}(T, P) 
\end{eqnarray}

\paragraph{Garment localizer} With the garment classes represented in a one hot encoded vector, we use the cross-entropy loss on the classification head and, for the regression head, we use the mean squared error.

\subsection{One landmark class per sample, constraint}
\label{subsec:spacial_contraint}
One important peculiarity of landmark detection to consider is the fact that, per image, only one class of landmarks exist. Therefore, we can introduce this constraint into the loss function and promote parameter combinations that tend to predict only one landmark per class. 
We implement this constraint by also applying cross-entropy over the spacial dimension, resulting in the loss function \ref{eq:spc_loss}. However, because cross-entropy expects its input to be a probability distribution, we must firstly  apply \textit{softmax} accordingly. We therefore, place two \textit{softmax} after the last convolutional layer activation: the first, (regular) depth wise, the second, spacial wise; and pass each output to the correspondent cross-entropy. At test time, we average the two \textit{softmax}, similarly to \ref{eq:spc_loss}. Because for each garment category, only a few landmarks are active, we further create a second mask, that is the ground-truth max spacial value, and we use it to ignore the loss spacial component for the landmark classes that are not applicable.

\begin{eqnarray}	 \label{eq:spc_loss}
	L_{cls} = \frac{1}{2} \Big[ m \odot L_{ce}(T, P) + \max_{w \times h} T \odot L_{ce(spatially)}(T, P) \Big]
\end{eqnarray}

\begin{figure}
  \begin{center}
    \leavevmode
    \includegraphics[width=1\textwidth]{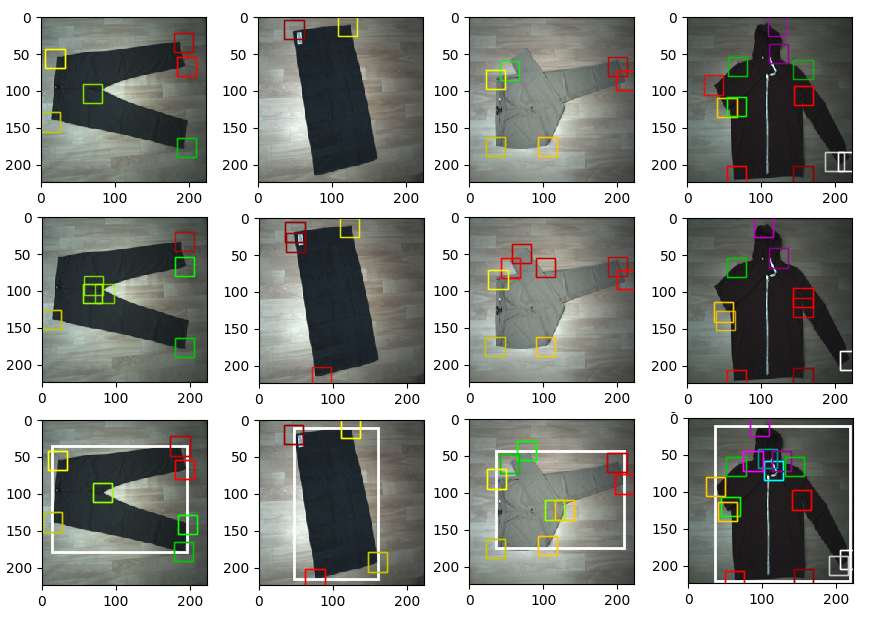}
    \caption{Representative cases of the result of applying the spacial constraint loss. At the top row, predictions with composed loss, at the middle, without, and the bottom, the ground truth. } 
    \label{fig:spacial_constraint}
  \end{center}
\end{figure}

Although with the spacial constraint loss addition, we achieve a 2\% lower detection mAP score, dropping from 37.8\% to 35.7\%. Yet, we are able to achieve lower duplicated predictions, as illustrated in \ref{fig:spacial_constraint}. 

\subsection{Using landmarks within garment localization} \label{subsec:bridge}
\begin{figure}
  \begin{center}
    \leavevmode
    \includegraphics[width=1\textwidth]{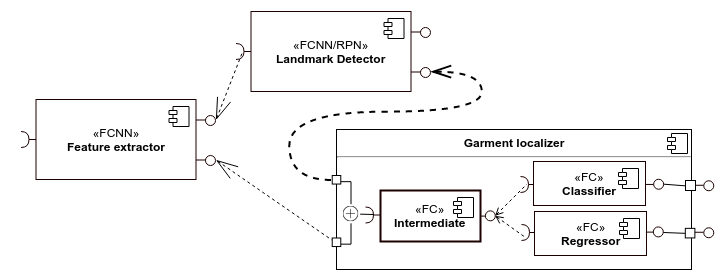}
    \caption{\textbf{GarmNet-B (introduced in \ref{subsec:bridge})} representation using UML\cite{uml}. The output emitted by the classifier block from the Landmark Detector branch is concatenated with the Feature Extractor output before being fed into the intermediate layer.} 
    \label{fig:bridge}
  \end{center}
\end{figure}


We investigate the gains in feeding the landmark detector output features into the garment localize intermediate layer, expecting that these would help better frame the garment bounding box and. We flatten the $18 \times 18$ tensor outputted by the classifier block from the landmark detector branch, and concatenate it with the flatten Feature Extractor output, before feeding it to the  512-d intermediate layer, resulting in GarmNet-B, represented in the figure \ref{fig:bridge}. With this bridge connection, the network achieves 32.0\% classification+localization error rate, a 24.7\% improvement when comparing with the individual garment detector training.

\begin{table}
\centering
\caption{Summary of Classification and Classification+Localization error rates. GarmNet is our base model, GarmNet-B is the model obtained after the bridge connection introduced in \ref{subsec:bridge} and GarmNet-B (A.D) the latter model optimized using the augmented dataset.}

\def\arraystretch{1.2}%
\begin{tabular}{l|l|l|}
\cline{2-3}
                                     & Classification & Classif.+Loca. \\ \hline
\multicolumn{1}{|l|}{GarmNet}        & 0\%     & 56.7\%         \\ \hline
\multicolumn{1}{|l|}{GarmNet-B}      & 0\%    & 32.0 \%   \\ \hline
\multicolumn{1}{|l|}{GarmNet-B (A.D.)} & 0\%          & 17.8\%   \\ \hline
\end{tabular}
\end{table}

\subsection{Final optimization with Augmented Data}

We perform one last training, without loading any previous learned parameters (with the exception of the feature extractor ImageNet parameters) and using augmented data. The data augmentation is achieved by repeating examples of less numerous classes and adding Gaussian and hue noise. The obtained results are: 0\% classification and 17.8\% classification+localization error rate, and 36.2\% landmark detection mAP. The complete classification accuracy can be justified by the almost constant background and the few, often differently colored, garments per class.  
\section{Conclusion}

In this work, we proposed a novel deep neural network model named GarmNet that can be optimized in an end-to-end manner, for performing simultaneous garment local and global perception. Approaches as ours are important for robotics applications, as these offer scalable to many classes, memory and processing efficient solutions, enabling real-time perception capabilities. 
We  evaluate  our  solution  using an augmented dataset assembled using the two collected by CTU during the CloPeMa project \cite{ctuDataset}. The experiments showed the effectiveness and side effects of introducing domain specific knowledge into the loss function being optimized, at both quantitative and qualitative levels. We finally demonstrate the improvements on garment localization by considering the landmark detection as an intermediate step.

In the future work, more experiments will be done to further optimise the network architecture and its hyper-parameters configuration. A more challenging dataset, with higher number of images and variability e.g., \cite{liu2016deepfashion}, will also be used. Within the context of garment perception for robotic laundry folding, the work will be extended to garment folding with a robot arm-hand setup, supported by the garment perception done in this paper and also possibly assisted by tactile sensors \cite{luo2017robotic,luo2018iclap,lee2019touching}.

\section*{ACKNOWLEDGMENT}
This work was supported by the EPSRC project ``Robotics and Artificial Intelligence for Nuclear (RAIN)" (EP/R026084/1).

\bibliographystyle{splncs04}
\bibliography{refs}
\clearpage
\section{Appendix}

\begin{table}
\def\arraystretch{1.1}%
\caption{Summary of landmark Classification+Localization, as follows: GarmNet, the results obtained using the base model; GarmNet (S.C), the base model optimized with the spacial constraint (\ref{subsec:spacial_contraint}); GarmNet-B, the modified model with the bridge connection (\ref{subsec:bridge}); and, GarmNet-B (A.D.), the latter model optimized using augmented dataset.}
\begin{tabular}{|l|l|l|l|l|}
\hline
                   & GarmNet & GarmNet (S.C.) & GarmNet-B & \begin{tabular}[c]{@{}l@{}}GarmNet-B (A.D.)\end{tabular} \\ \hline
mean AP            & 37.8    & 35.7      & 34.5      & 36.1                                                        \\ \hline
Left leg outer     & 38.5    & 33.7      & 33.0      & 38.2                                                        \\ \hline
Left leg inner     & 47.6    & 39.5      & 0         & 44.1                                                        \\ \hline
Crotch             & 46.4    & 42.5      & 43.7      & 34.3                                                        \\ \hline
Right leg inner    & 46.0    & 42.6      & 50.5      & 39.9                                                        \\ \hline
Left leg inner     & 45.3    & 41.4      & 43.7      & 40.9                                                        \\ \hline
Top right          & 62.1    & 54.7      & 50.5      & 55.7                                                        \\ \hline
Top left           & 57.3    & 58.4      & 47.8      & 50                                                          \\ \hline
Right sleave inner & 44.3    & 38.7      & 60.2      & 37.3                                                        \\ \hline
Right sleave outer & 38.2    & 30.4      & 53.6      & 42.2                                                        \\ \hline
Left sleave inner  & 43.9    & 41.5      & 46.5      & 48.3                                                        \\ \hline
Left selave outer  & 34.5    & 30.7      & 46.0      & 35.7                                                        \\ \hline
Hood right         & 0       & 54.5      & 46.4      & 0                                                           \\ \hline
Hood top           & 0       & 0         & 42.5      & 40.9                                                        \\ \hline
Hood left          & 56.4    & 30.9      & 0         & 0                                                           \\ \hline
Bottom left        & 38.8    & 38.2      & 34.5      & 40.7                                                        \\ \hline
Bottom middle      & 0       & 0         & 0         & 43.9                                                        \\ \hline
Bottom right       & 38.6    & 37.1      & 40        & 39.9                                                        \\ \hline
Right armpit       & 46.1    & 41.7      & 44.7      & 50.7                                                        \\ \hline
Right shoulder     & 44.0    & 39.7      & 47.1      & 60.2                                                        \\ \hline
Neckline right     & 44.5    & 32.2      & 45.0      & 36.1                                                        \\ \hline
Collar right       & 44.1    & 38.7      & 47.2      & 48.7                                                        \\ \hline
Collar left        & 55.3    & 48.4      & 51.9      & 60.2                                                        \\ \hline
Neckline left      & 34.3    & 33.8      & 35.6      & 36.1                                                        \\ \hline
Left shoulder      & 44.4    & 39.1      & 46.7      & 48.7                                                        \\ \hline
Left armpit        & 43.3    & 38.8      & 39.0      & 37.4                                                        \\ \hline
Fold 1             & 27.5    & 26.4      & 27.5      & 25.2                                                        \\ \hline
Fold 2             & 0       & 0         & 0         & 0                                                           \\ \hline
\end{tabular}
\end{table}

\end{document}